%% file: emnlp2020.tex
\documentclass[11pt,a4paper]{article}
\usepackage{xr-hyper}
\usepackage[hyperref]{emnlp2020}
\usepackage{times}
\usepackage{latexsym}
\usepackage{xspace}

\usepackage{graphicx}
\usepackage{float}
\usepackage{subcaption}
\usepackage{csquotes}
\usepackage{calc}
\usepackage{booktabs}
\usepackage{amsmath}
\usepackage{amssymb}
\usepackage{bbm}
\usepackage{dblfloatfix}
\usepackage[most]{tcolorbox}
\usepackage{subfiles}
\usepackage{balance}
\graphicspath{{./}{../}}

\definecolor{cameraready}{HTML}{000000}
\definecolor{correct}{HTML}{0571B0}
\definecolor{incorrect}{HTML}{F4A582}

\usepackage{microtype}

\aclfinalcopy 

\title{Domain-Specific Lexical Grounding in Noisy Visual-Textual Documents}

\author{Gregory Yauney \\
  Cornell University \\
  \texttt{\small{gyauney@cs.cornell.edu}} \\\And
  Jack Hessel \\
  Allen Institute for AI \\
  \texttt{\small{jackh@allenai.org}} \\\And
  David Mimno \\
  Cornell University \\
  \texttt{\small{mimno@cornell.edu}} \\}

\date{November 2020}

\include{macros}

\begin{document}
\maketitle
\begin{abstract}
Images can give us insights into the contextual meanings of words, but
current image-text grounding approaches require detailed annotations.
Such granular annotation is rare, expensive, and unavailable in most domain-specific contexts. 
In contrast, unlabeled multi-image, multi-sentence documents are abundant.
Can lexical grounding be learned from such documents, even though they have significant lexical and visual overlap?
Working with a case study dataset of real estate listings, we demonstrate the challenge of distinguishing highly correlated grounded terms, such as ``kitchen'' and ``bedroom'', and introduce metrics to assess this document similarity.
We present a simple unsupervised clustering-based method that increases precision and recall beyond object detection and image tagging baselines when evaluated on labeled subsets of the dataset.
The proposed method is particularly effective for local contextual meanings of a word, for example associating ``granite'' with countertops in the real estate dataset and with rocky landscapes in a Wikipedia dataset.
\end{abstract}

\subfile{sections/introduction}



\subfile{sections/models}

\subfile{sections/results-panel}

\subfile{sections/experiments}

\subfile{sections/discussion}

\section*{Acknowledgments}
We would particularly like to thank Grant Long for putting together the StreetEasy data
as well as \mbox{Ondrej} Linda, Ramin Mehran, and Randy Puttick for helpful conversations.
We would like to thank Maria Antoniak for valuable feedback.
Data provided by StreetEasy, an affiliate of Zillow Group. The results and opinions are those of the authors and do not reflect the position of StreetEasy, Zillow Group, or any of their affiliates.
This work was supported by Zillow Group and NSF \#1652536. JH performed this work while at Cornell University.

\balance
\bibliography{emnlp2020}
\bibliographystyle{acl_natbib}

\appendix

\subfile{sections/appendix}

\end{document}

%% file: macros.tex
\newif\iffinal

\finalfalse
\iffinal
\newcommand{\jack}[1]{}
\else
\newcommand{\jack}[1]{\textcolor{blue}{\textbf{Jack: #1}}}
\fi

\newcommand{\algname}{EntSharp\xspace}

%% file: sections/introduction.tex
\begin{figure}[t!]
	\scriptsize
	\sffamily
	\centering
	\textbf{\enquote{granite}}\\[1mm]
	\begin{subfigure}[c]{.98\linewidth}
		\centering  
		\parbox{\widthof{StreetEasyM}}{
			\raisebox{\dimexpr 8mm}{StreetEasy}
		}
		\includegraphics[height=9mm]{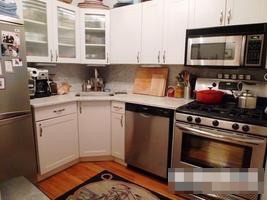}\hfill
		\includegraphics[height=9mm]{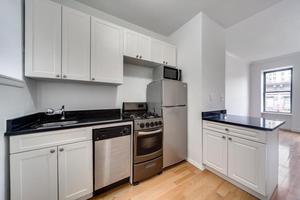}\hfill
		\includegraphics[height=9mm]{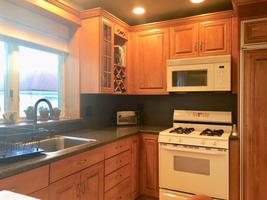}\hfill
		\includegraphics[height=9mm]{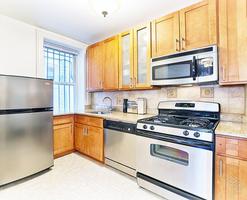}
  	\end{subfigure}\\[-3mm]

	\begin{subfigure}[c]{.98\linewidth}
		\centering  
		\parbox{\widthof{StreetEasyM}}{
			\raisebox{\dimexpr 8mm}{Wikipedia}
		}
		\includegraphics[height=11mm]{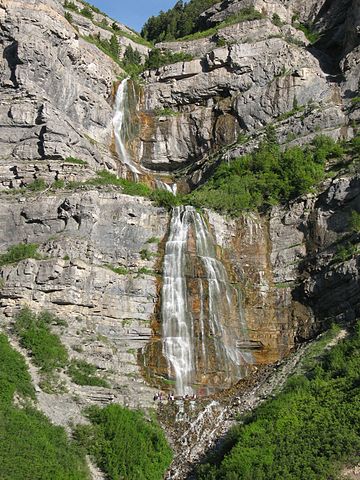}\hfill
		\includegraphics[height=11mm]{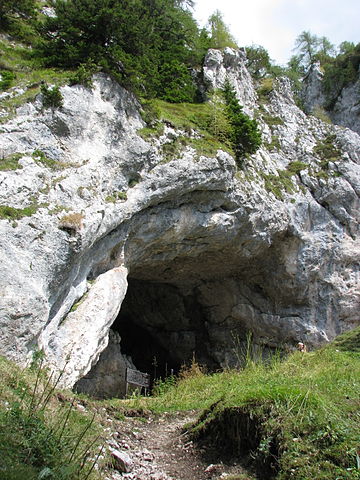}\hfill
		\includegraphics[height=11mm]{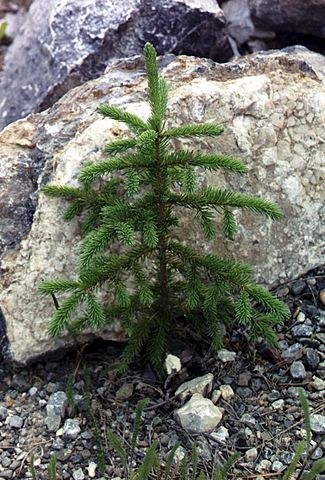}\hfill
		\includegraphics[height=11mm]{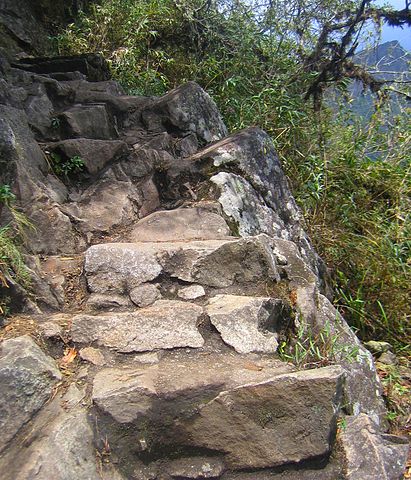}\hfill
		\includegraphics[height=11mm]{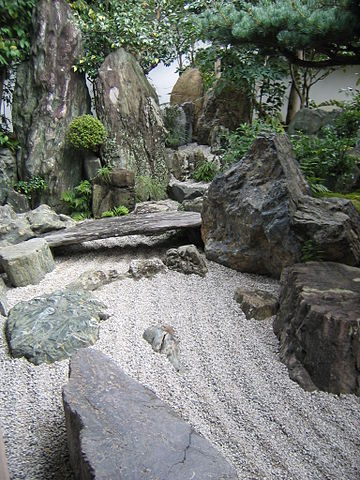}\hfill
		\includegraphics[height=11mm]{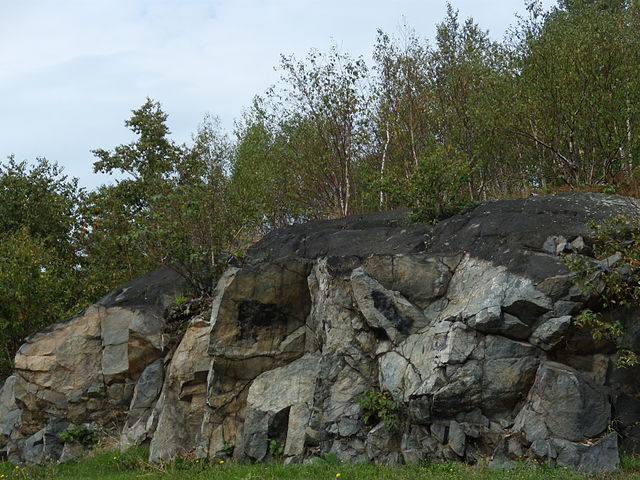}
	\end{subfigure}

	\textbf{\enquote{chrysler}}\\[1mm]
	\begin{subfigure}[c]{.98\linewidth}
		\centering  
		\parbox{\widthof{StreetEasyM}}{
			\raisebox{\dimexpr 8mm}{StreetEasy}
		}
		\includegraphics[height=9mm]{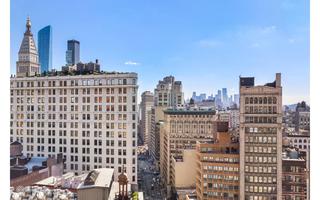}\hfill
		\includegraphics[height=9mm]{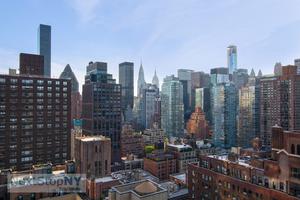}\hfill
		\includegraphics[height=9mm]{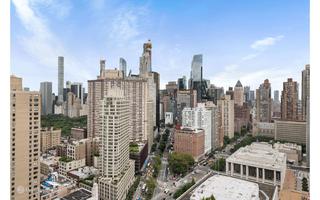}\hfill
		\includegraphics[height=9mm]{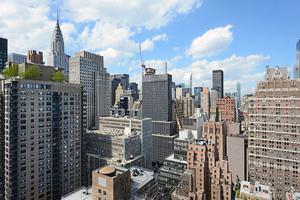}
  	\end{subfigure}\\[-3mm]

	\begin{subfigure}[c]{.98\linewidth}
		\centering  
		\parbox{\widthof{StreetEasyM}}{
			\raisebox{\dimexpr 8mm}{Wikipedia}
		}
		\includegraphics[height=9mm]{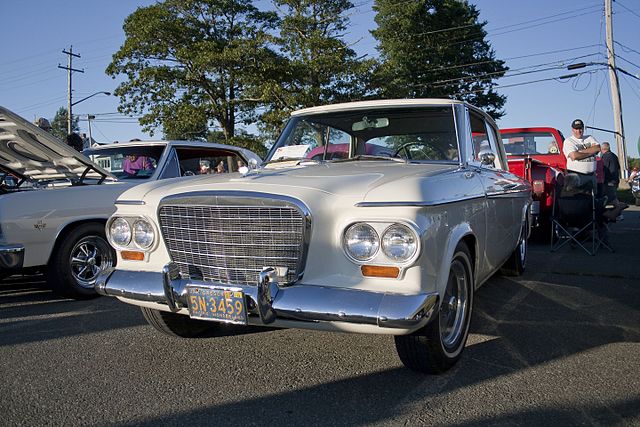}\hfill
		\includegraphics[height=9mm]{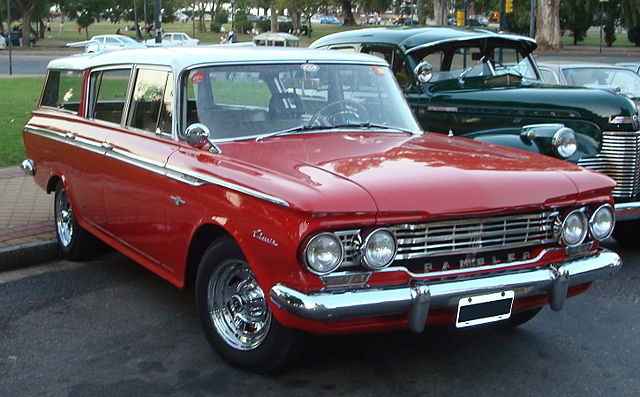}\hfill
		\includegraphics[height=9mm]{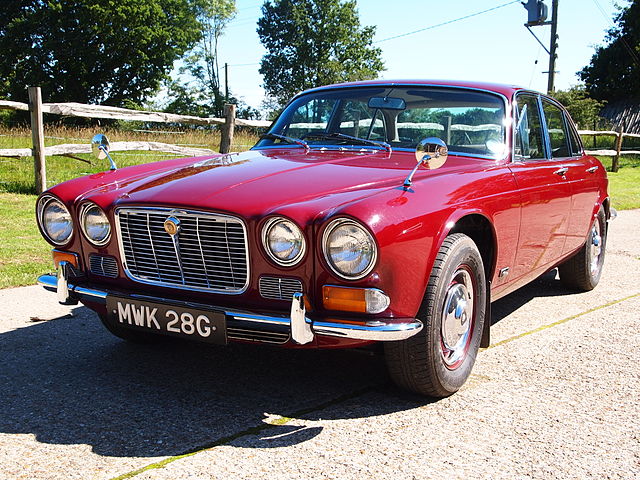}\hfill
		\includegraphics[height=9mm]{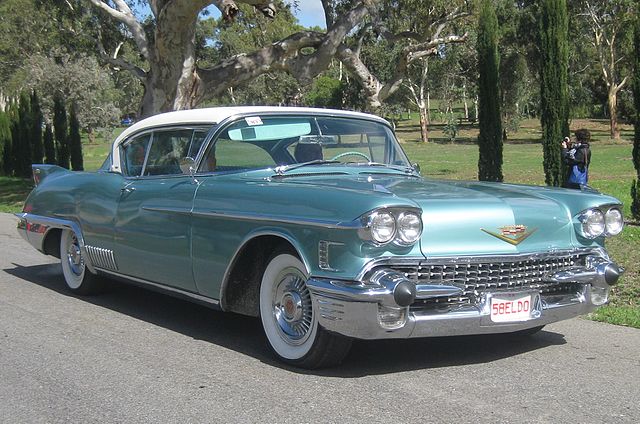}
	\end{subfigure}

	\textbf{\enquote{architect}}\\[1mm]
	\begin{subfigure}[c]{.98\linewidth}
		\centering  
		\parbox{\widthof{StreetEasyM}}{
			\raisebox{\dimexpr 8mm}{StreetEasy}
		}
		\includegraphics[height=9mm]{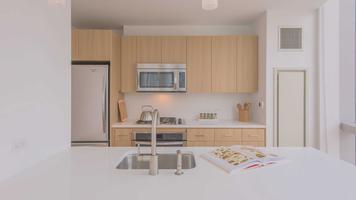}\hfill
		\includegraphics[height=9mm]{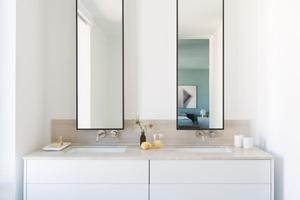}\hfill
		\includegraphics[height=9mm]{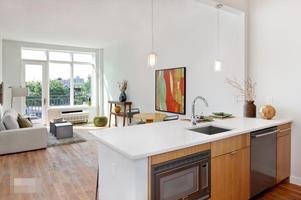}\hfill
		\includegraphics[height=9mm]{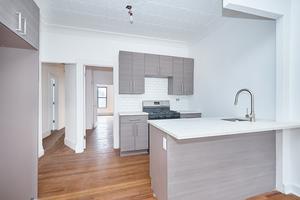}
  	\end{subfigure}\\[-3mm]

	\begin{subfigure}[c]{.98\linewidth}
		\centering  
		\parbox{\widthof{StreetEasyM}}{
			\raisebox{\dimexpr 8mm}{Wikipedia}
		}
		\includegraphics[height=9mm]{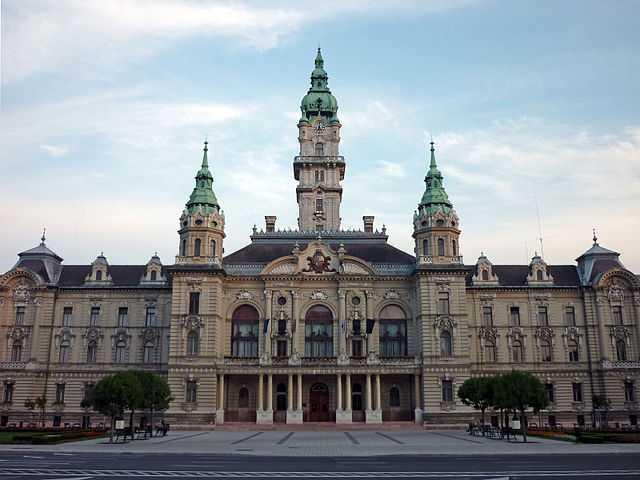}\hfill
		\includegraphics[height=9mm]{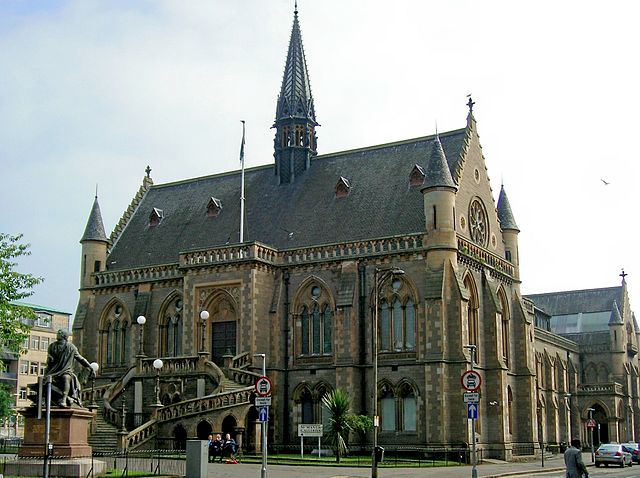}\hfill
		\includegraphics[height=9mm]{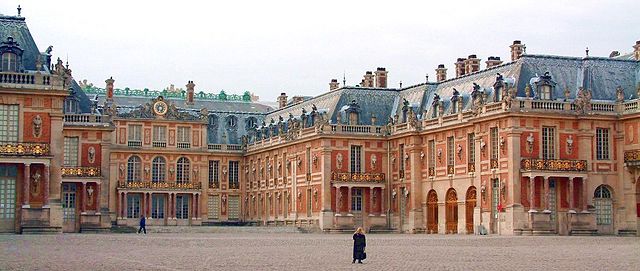}\hfill
		\includegraphics[height=9mm]{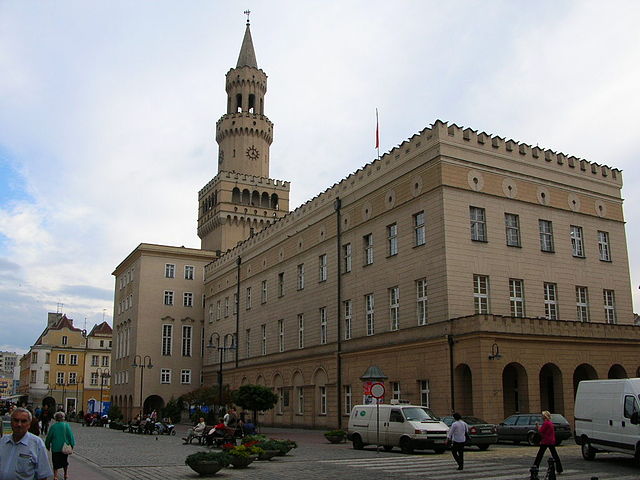}
	\end{subfigure}

  \caption{We identify domain-specific associations between words and images from unlabeled multi-sentence, multi-image documents.}
  \label{fig:comparison}
\end{figure}

\section{Introduction}
\label{sec:sec_with_identifiability}

\begin{figure*}[t!]
  \centering
    \includegraphics[width=\linewidth]{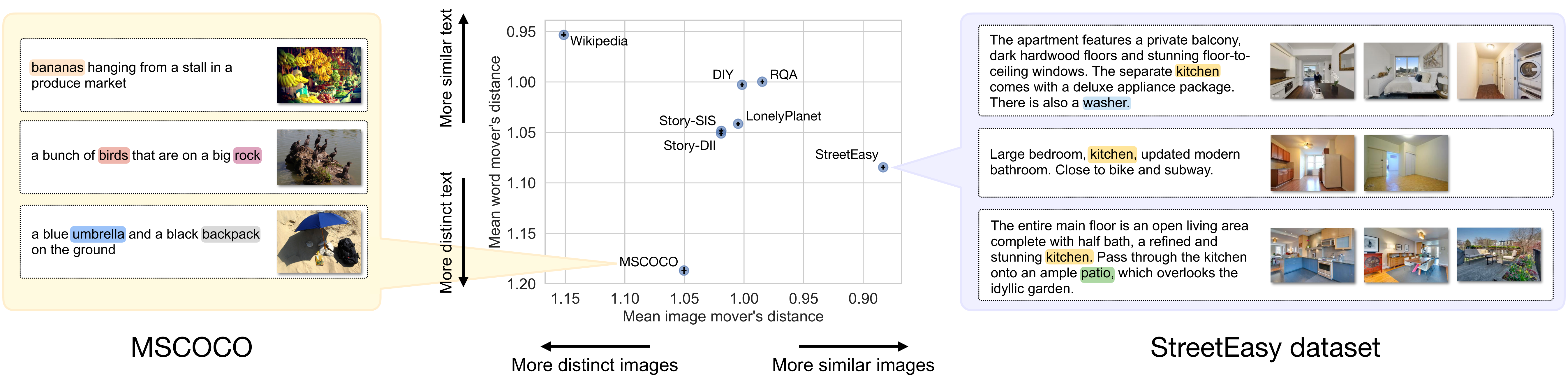}
  \caption{Documents in the StreetEasy dataset are much more visually similar to each other than documents in seven multimodal image-text datasets spanning storytelling, cooking, travel blogs, captioning, etc. \cite{lin2014microsoft,huang2016visual,yagcioglu2018recipeqa,hessel2018quantifying,hessel2019unsupervised,nag2020illustrate}. Examples from StreetEasy show that words like \enquote{kitchen} are frequent and grounded. Black lines represent 99.99\% CI.}
  \label{fig:overview}
\end{figure*}

Multimodal data consisting of text and images is not only ubiquitous
but increasingly diverse: libraries are digitizing visual-textual
collections \cite{British_Library_Labs_2016,smithsoniandata}; news
organizations release over 1M images per year to accompany news
articles \cite{associatedpress}; and social media messages are rarely
sent without visual accompaniment.  In this work, we focus on one such
specialized, multimodal domain: New York City real estate listings from the website StreetEasy.

To effectively index image-text datasets for search, retrieval,
and other tasks, we need algorithms that learn connections between
modalities, doing so from data that is naturally abundant.  In
{\color{cameraready} documents that contain multiple images and sentences}, there may be \emph{no explicit annotations} for
image-sentence associations or bounding box-word associations. As a result, existing
{\color{cameraready} image captioning/tagging methods are difficult to adapt to \emph{unlabeled} multi-image, multi-sentence documents.}
Indeed, most prior image captioning work
has focused on rare and expensive
single-image, single-caption collections such as MSCOCO, which focuses on literal, context-free descriptions
for 80 object types \cite{lin2014microsoft}. Similarly, off-the-shelf object detectors may not
account for contextual factors: to an ImageNet classifier, ``pool''
refers to a pool table \cite{russakovsky2015imagenet}. In the specialized real estate context, ``pool'' commonly refers to a swimming pool.



{\color{cameraready}Consider the task of lexical grounding: given a word, which images in the corpus depict that word?}
Consider the difficulty in learning a visual grounding for ``kitchen'' in StreetEasy.
First, documents are multi-image, multi-sentence rather than single-image,
single-sentence.
Second, almost all documents picture a
kitchen, a living room, and a dining room.
Finally,
\enquote{kitchen} co-occurs with more than two-thirds of all images,
the majority of which are not kitchens. Is this task even possible?

{\color{cameraready} Our first contribution is to map out a landscape of multimodal
datasets, placing our real estate case-study in relation to existing
corpora.}
We operationalize this notion in Figure~\ref{fig:overview} by plotting
average across-document visual+textual similarity for our StreetEasy
case study compared to several existing multimodal
corpora;\footnote{We compute text similarity
  between documents with a length-controlled version of word mover's
  distance (WMD) \cite{kusner2015word} on word2vec token features. We compute visual similarity between documents with ``image  mover's'' distance, which is identical to WMD, but with a CNN feature for each
  image. More details are given in Appendix~\ref{app:similarity}.}
indeed, images in StreetEasy have very low 
diversity compared to other corpora. As a result of this self-similarity, in
\S\ref{sec:sec_with_results}, we find that image-text grounding is
difficult for off-the-shelf image tagging methods like
multinomial/softmax regression, which leverage variation in both
lexical and visual features across documents.\footnote{Existing
  unsupervised approaches for this setting
  \cite{hessel2019unsupervised,nag2020illustrate} learn within-document matchings of whole sentences/paragraphs, we learn cross-document matchings of word types to images.}






Our second contribution is a simple but performant clustering
algorithm for this setting, \emph{\algname}.\footnote{Code is at \url{https://github.com/gyauney/domain-specific-lexical-grounding}.} We intend this method to learn from $\langle \textrm{image}, \textrm{word} \rangle$
co-occurrences collected from multi-image, multi-sentence document collections. The training process iteratively ``sharpens'' the
estimated $\Pr(\textrm{word} \, |\, \textrm{image})$ distributions so
that words ``compete'' to claim responsibility for images.
We show that \algname outperforms both object
detection and image tagging baselines at retrieving relevant images
for given word types. We then qualitatively explore \algname's
predictions on both StreetEasy and a multimodal Wikipedia dataset
\cite{hessel2018quantifying}. The algorithm is often able to learn
corpus specific relations: as shown in Figure \ref{fig:comparison}, in
the context of NYC real estate, \enquote{chrysler} refers to a
prominent building and \enquote{granite} to a kitchen surface, while in
Wikipedia the same words are grounded in cars and rocky outcroppings.

\paragraph{Related work.} Learning image-text relationships
is central to many applications, including image captioning/tagging
\cite{kulkarni2013babytalk,mitchell2013generating,karpathy2015deep} and cross-modal
retrieval/search \cite{jeon2003automatic,rasiwasia2010new}. While
most captioning work assumes a supervised one-to-one corpus, recent works consider
documents containing multiple images/sentences
\cite{park2015expressing,shin2016beyond,agrawal2016sort,liu2017let,chu2017blog,hessel2019unsupervised,nag2020illustrate}. Furthermore, compared
to crowd-annotated captioning datasets, web corpora are more challenging, as
image-text relationships often transcend literal description
\cite{marsh2003taxonomy,alikhani2019caption}.

%% file: sections/models.tex
\section{Task and Models}

We consider a direct image-text grounding task: for each word type, we
aim to retrieve images most-associated with that word. Models are
evaluated by their capacity to compute word-image similarities
that align with human judgment.

\paragraph{\algname.}
For each image in a document we iteratively infer a probability distribution over the words present in the document. 
During training, these distributions are encouraged to have low entropy.
The output is an embedding of each word into image space: the model computes word-image similarities in this joint space.
This can be thought of as a soft clustering, such that each word type is equivalent to a cluster but only certain clusters are available to certain images.
{\color{cameraready}
This approach could also be situated within the framework of multiple-instance learning \cite{carbonneau2018multiple}.}


Each image $i$ starts with a fixed feature vector $\vec{i} \in \mathbb{R}^d$. Let $\mathcal{I}$ be the set of these image embeddings. For each word $w$ we initialize a cluster centroid $\vec{w} \in \mathbb{R}^d$  to the average of co-occurring images' embeddings.
Let $\mathbbm{1}_{i,w}$ be 1 if image $i$ co-occurs with word $w$ in any document and 0 otherwise.
Each image $\vec{i}$ is assumed to have a membership distribution $\vec{p}_i$ over words, where $\vec{p}_i$ is initially uniform over co-occurring words.
At each iteration, cluster centroids are updated to the weighted average of co-occurring images' embeddings: $\vec{w} := \sum_{\vec{i} \in \mathcal{I}} p_i(w) \cdot \vec{i}$ followed by normalization.
Each image's distribution over clusters is updated by taking a softmax of the cosine similarity between pairs of image and word embeddings, first multiplying similarities by a sharpness coefficient\footnote{Sharpness is equal to the inverse of softmax temperature; thus \algname equivalently decreases softmax temperature during training.} equal to the iteration number, and finally masking for co-occurrence:
$p_i(w) \propto \mathbbm{1}_{i,w} \cdot \exp \big( \text{sharpness} \cdot ( \vec{i} \cdot \vec{w} ) \big)$.
After training, we calculate the cosine similarity between image embeddings and the learned word-cluster embedding.

\paragraph{Untrained \algname baseline.}
We consider a simple averaging baseline, corresponding to the cluster center initializations of \algname: each word embedding is set to the mean of the features for all its co-occurring images.

\paragraph{Object detection baselines.}
{\color{cameraready}
We can use ImageNet to identify objects, but most words in the full vocabulary are not in the ImageNet labels.
}
We implement two object detection baselines that map images to object names and then match object names to words in documents \cite{hessel2019unsupervised}.
{\color{cameraready}
For each image, we first get the image's top class predictions from DenseNet169 \cite{huang2017densely} pretrained on the ImageNet classification task \cite{russakovsky2015imagenet}. These predictions are for a whole image and are restricted to the 1000 ImageNet labels.
We bridge the gap between ImageNet labels and the vocabulary by then creating an \emph{image vector} by averaging the word vectors corresponding to these predictions.
Finally, for each word in the full vocabulary, we rank images by the cosine similarity between the word's vector and these image vectors.
Words are represented in one baseline by word2vec embeddings \cite{mikolov2013distributed} and in the other by the output of RoBERTa \cite{liu2019roberta} when fed a single token as input.
}

\paragraph{Image tagging baselines.} Inspired by \newcite{mahajan2018exploring},
we implement softmax and multinomial regression models.
{\color{cameraready}
The former, softmax regression, takes image features and predicts a distribution over the words in the vocabulary with a softmax loss.
It computes the word type indicator vector for each document, i.e., 1 if word $w$ was in the document else 0, and then $\ell_1$ normalizes.
Multinomial regression computes the word type indicator vector, and---instead of normalizing---computes the logistic sigmoid loss treating the labels as 0/1 indicators.
This is equivalent to training a separate logistic regression for each word type to predict the presence/absence of a word type in each document, given the image features.
Both models finally use the predicted conditional distributions to produce a ranking of images for each word.
}


%% file: sections/results-panel.tex
\begin{figure*}[t!]
	\sffamily
	\setlength{\fboxsep}{1.25pt}\setlength{\fboxrule}{0pt}%
	
	\begin{subfigure}[c]{0.48\textwidth}
	\scriptsize
	\textbf{\enquote{kitchen}} (18.4\% labeled true)\\[-2mm]
  
	\parbox{\widthof{\scriptsize{W: 88.8 \textsf{\textsc{auc}}}}}{
		\raisebox{\dimexpr 6mm}{\scriptsize{E: 72.9 \textsf{\textsc{auc}}}}
	}
	\fcolorbox{white}{correct}{\includegraphics[height=7mm]{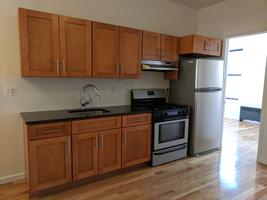}}\hfill
	\fcolorbox{white}{correct}{\includegraphics[height=7mm]{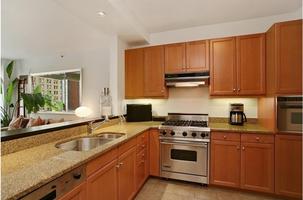}}\hfill
	\fcolorbox{white}{correct}{\includegraphics[height=7mm]{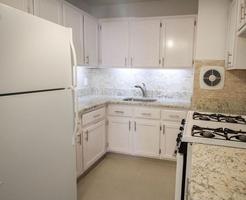}}\hfill
	\fcolorbox{white}{correct}{\includegraphics[height=7mm]{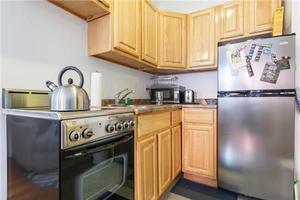}}\hfill
	\fcolorbox{white}{correct}{\includegraphics[height=7mm]{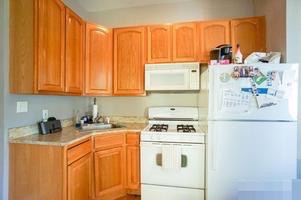}}
	\\[-2mm]

	\parbox{\widthof{\scriptsize{W: 88.8 \textsf{\textsc{auc}}}}}{
		\raisebox{\dimexpr 6mm}{\scriptsize{W: 52.7 \textsf{\textsc{auc}}}}
	}
  	\fcolorbox{white}{incorrect}{\includegraphics[height=7mm]{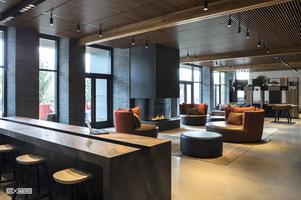}}\hfill
	\fcolorbox{white}{correct}{\includegraphics[height=7mm]{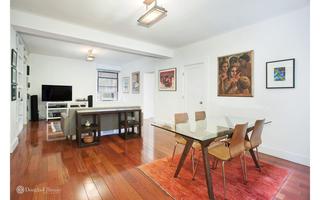}}\hfill
	\fcolorbox{white}{correct}{\includegraphics[height=7mm]{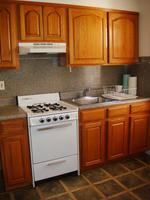}}\hfill
	\fcolorbox{white}{correct}{\includegraphics[height=7mm]{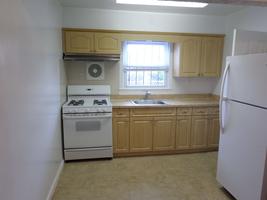}}\hfill
	\fcolorbox{white}{correct}{\includegraphics[height=7mm]{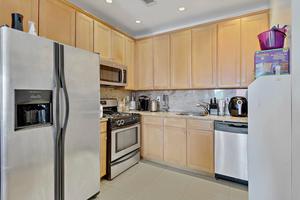}}
	\\[-2mm]

	\parbox{\widthof{\scriptsize{W: 88.8 \textsf{\textsc{auc}}}}}{
		\raisebox{\dimexpr 6mm}{\scriptsize{R: 21.1 \textsf{\textsc{auc}}}}
	}
  \fcolorbox{white}{correct}{\includegraphics[height=7mm]{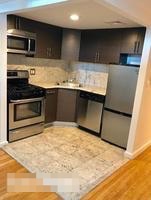}}\hfill
	\fcolorbox{white}{correct}{\includegraphics[height=7mm]{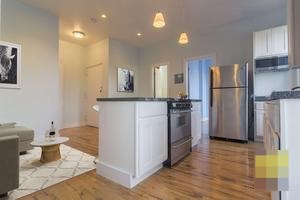}}\hfill
	\fcolorbox{white}{correct}{\includegraphics[height=7mm]{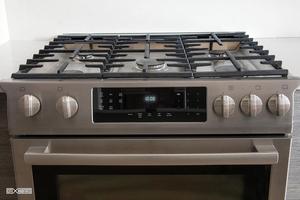}}\hfill
	\fcolorbox{white}{correct}{\includegraphics[height=7mm]{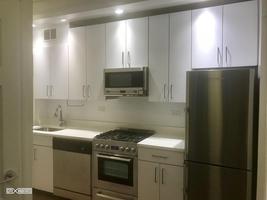}}\hfill
	\fcolorbox{white}{correct}{\includegraphics[height=7mm]{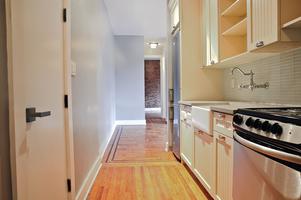}}
  \end{subfigure}
  \hfill
  \begin{subfigure}[c]{0.48\textwidth}
	\scriptsize
	\textbf{\enquote{outdoor}} (16.9\% labeled true)\\[-2mm]
  
	\parbox{\widthof{\scriptsize{W: 88.8 \textsf{\textsc{auc}}}}}{
		\raisebox{\dimexpr 6mm}{\scriptsize{E: 68.5 \textsf{\textsc{auc}}}}
	}
	\fcolorbox{white}{correct}{\includegraphics[height=7mm]{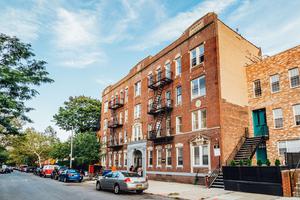}}\hfill
	\fcolorbox{white}{correct}{\includegraphics[height=7mm]{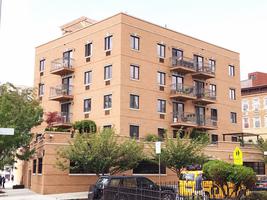}}\hfill
	\fcolorbox{white}{correct}{\includegraphics[height=7mm]{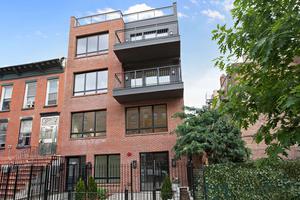}}\hfill
	\fcolorbox{white}{correct}{\includegraphics[height=7mm]{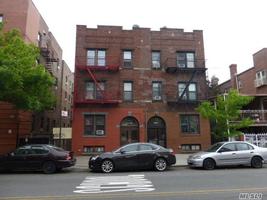}}\hfill
	\fcolorbox{white}{correct}{\includegraphics[height=7mm]{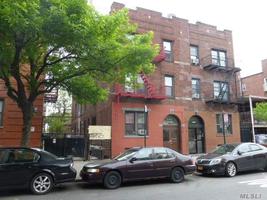}}
	\\[-2mm]

	\parbox{\widthof{\scriptsize{W: 88.8 \textsf{\textsc{auc}}}}}{
		\raisebox{\dimexpr 6mm}{\scriptsize{W: 20.0 \textsf{\textsc{auc}}}}
	}
  	\fcolorbox{white}{incorrect}{\includegraphics[height=7mm]{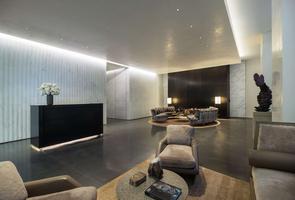}}\hfill
	\fcolorbox{white}{incorrect}{\includegraphics[height=7mm]{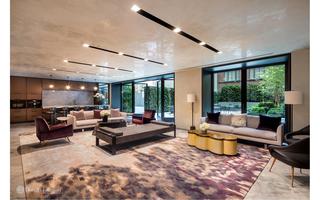}}\hfill
	\fcolorbox{white}{correct}{\includegraphics[height=7mm]{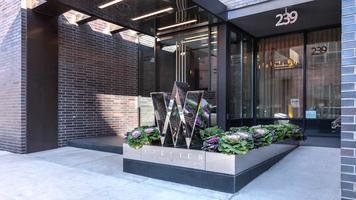}}\hfill
	\fcolorbox{white}{correct}{\includegraphics[height=7mm]{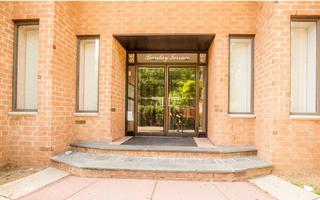}}\hfill
	\fcolorbox{white}{correct}{\includegraphics[height=7mm]{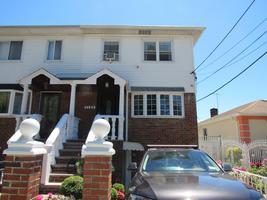}}
	\\[-2mm]

	\parbox{\widthof{\scriptsize{W: 88.8 \textsf{\textsc{auc}}}}}{
		\raisebox{\dimexpr 6mm}{\scriptsize{R: 13.2 \textsf{\textsc{auc}}}}
	}
  \fcolorbox{white}{correct}{\includegraphics[height=7mm]{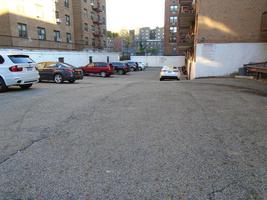}}\hfill
	\fcolorbox{white}{incorrect}{\includegraphics[height=7mm]{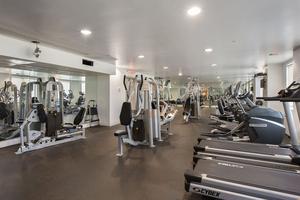}}\hfill
	\fcolorbox{white}{incorrect}{\includegraphics[height=7mm]{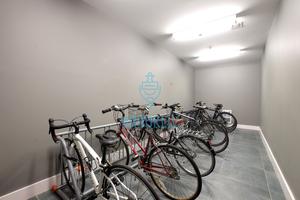}}\hfill
	\fcolorbox{white}{incorrect}{\includegraphics[height=7mm]{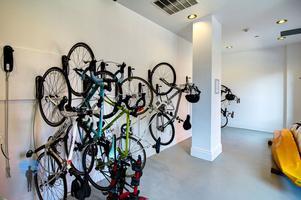}}\hfill
	\fcolorbox{white}{incorrect}{\includegraphics[height=7mm]{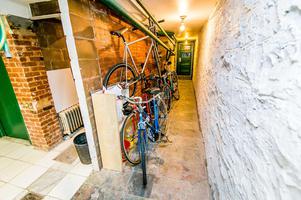}}
  \end{subfigure}

  \begin{subfigure}[c]{0.48\textwidth}
	\scriptsize
	\textbf{\enquote{washer}} (1.6\% labeled true)\\[-2mm]
  
	\parbox{\widthof{\scriptsize{W: 88.8 \textsf{\textsc{auc}}}}}{
		\raisebox{\dimexpr 6mm}{\scriptsize{E: 70.7 \textsf{\textsc{auc}}}}
	}
	\fcolorbox{white}{correct}{\includegraphics[height=7mm]{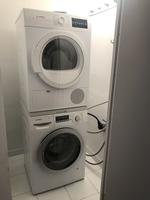}}\hfill
	\fcolorbox{white}{correct}{\includegraphics[height=7mm]{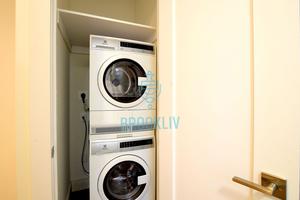}}\hfill
	\fcolorbox{white}{correct}{\includegraphics[height=7mm]{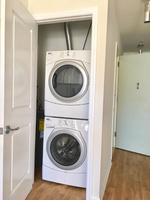}}\hfill
	\fcolorbox{white}{correct}{\includegraphics[height=7mm]{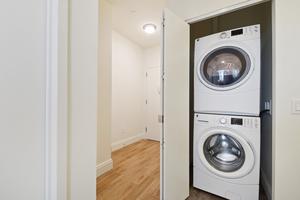}}\hfill
	\fcolorbox{white}{correct}{\includegraphics[height=7mm]{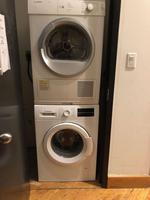}}
	\\[-2mm]

	\parbox{\widthof{\scriptsize{W: 88.8 \textsf{\textsc{auc}}}}}{
		\raisebox{\dimexpr 6mm}{\scriptsize{W: 49.3 \textsf{\textsc{auc}}}}
	}
  	\fcolorbox{white}{correct}{\includegraphics[height=7mm]{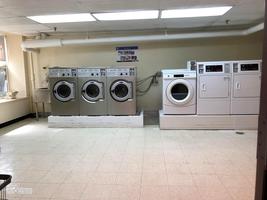}}\hfill
	\fcolorbox{white}{correct}{\includegraphics[height=7mm]{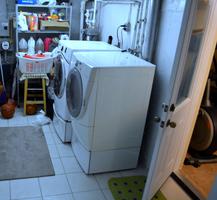}}\hfill
	\fcolorbox{white}{correct}{\includegraphics[height=7mm]{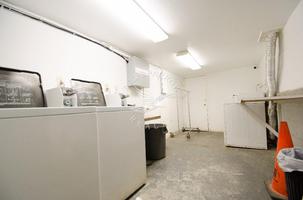}}\hfill
	\fcolorbox{white}{correct}{\includegraphics[height=7mm]{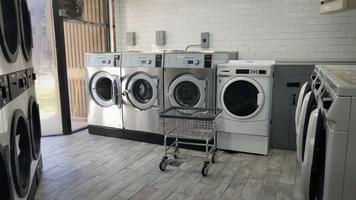}}\hfill
	\fcolorbox{white}{correct}{\includegraphics[height=7mm]{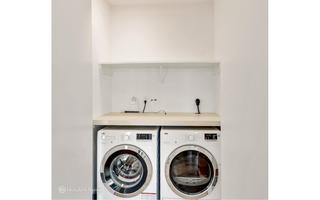}}
	\\[-2mm]

	\parbox{\widthof{\scriptsize{W: 88.8 \textsf{\textsc{auc}}}}}{
		\raisebox{\dimexpr 6mm}{\scriptsize{R: 62.1 \textsf{\textsc{auc}}}}
	}
  \fcolorbox{white}{correct}{\includegraphics[height=7mm]{figures/350992643.jpg}}\hfill
	\fcolorbox{white}{correct}{\includegraphics[height=7mm]{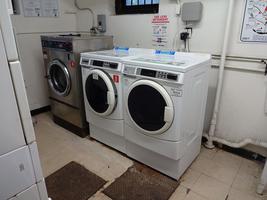}}\hfill
	\fcolorbox{white}{correct}{\includegraphics[height=7mm]{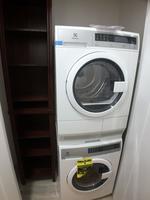}}\hfill
	\fcolorbox{white}{correct}{\includegraphics[height=7mm]{figures/336850166.jpg}}\hfill
	\fcolorbox{white}{correct}{\includegraphics[height=7mm]{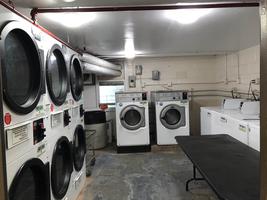}}
  \end{subfigure}
  \hfill
  \begin{subfigure}[c]{0.48\textwidth}
	\scriptsize
	\textbf{\enquote{pool}} (1.3\% labeled true)\\[-2mm]
  
	\parbox{\widthof{\scriptsize{W: 88.8 \textsf{\textsc{auc}}}}}{
		\raisebox{\dimexpr 6mm}{\scriptsize{E: 49.6 \textsf{\textsc{auc}}}}
	}
	\fcolorbox{white}{correct}{\includegraphics[height=7mm]{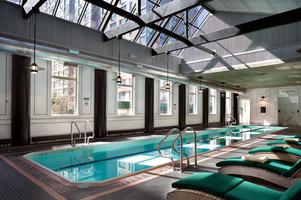}}\hfill
	\fcolorbox{white}{correct}{\includegraphics[height=7mm]{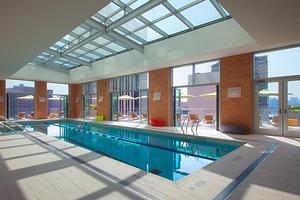}}\hfill
	\fcolorbox{white}{correct}{\includegraphics[height=7mm]{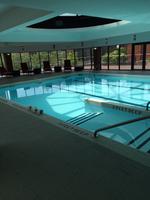}}\hfill
	\fcolorbox{white}{correct}{\includegraphics[height=7mm]{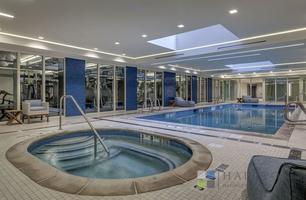}}\hfill
	\fcolorbox{white}{correct}{\includegraphics[height=7mm]{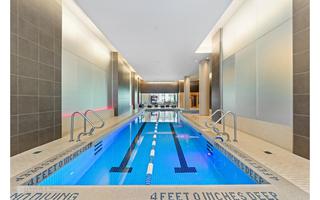}}
	\\[-2mm]

	\parbox{\widthof{\scriptsize{W: 88.8 \textsf{\textsc{auc}}}}}{
		\raisebox{\dimexpr 6mm}{\scriptsize{W: 20.1 \textsf{\textsc{auc}}}}
	}
  	\fcolorbox{white}{correct}{\includegraphics[height=7mm]{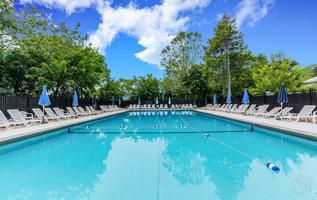}}\hfill
	\fcolorbox{white}{correct}{\includegraphics[height=7mm]{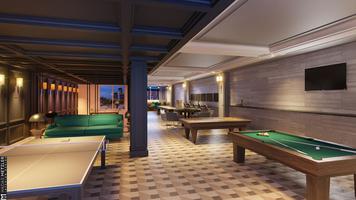}}\hfill
	\fcolorbox{white}{incorrect}{\includegraphics[height=7mm]{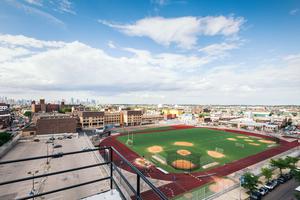}}\hfill
	\fcolorbox{white}{correct}{\includegraphics[height=7mm]{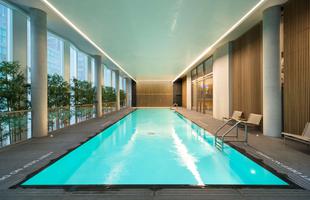}}\hfill
	\fcolorbox{white}{correct}{\includegraphics[height=7mm]{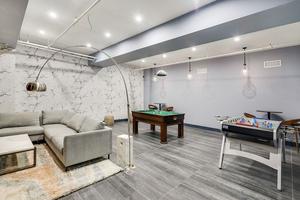}}
	\\[-2mm]

	\parbox{\widthof{\scriptsize{W: 88.8 \textsf{\textsc{auc}}}}}{
		\raisebox{\dimexpr 6mm}{\scriptsize{R: 17.2ß \textsf{\textsc{auc}}}}
	}
  \fcolorbox{white}{correct}{\includegraphics[height=7mm]{figures/346500513.jpg}}\hfill
	\fcolorbox{white}{correct}{\includegraphics[height=7mm]{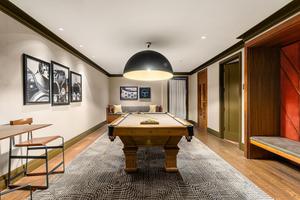}}\hfill
	\fcolorbox{white}{correct}{\includegraphics[height=7mm]{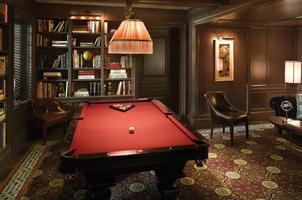}}\hfill
	\fcolorbox{white}{correct}{\includegraphics[height=7mm]{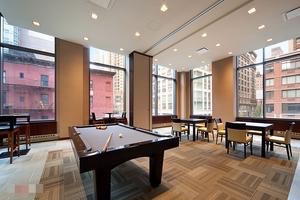}}\hfill
	\fcolorbox{white}{correct}{\includegraphics[height=7mm]{figures/331868835.jpg}}
  \end{subfigure}
  
  \begin{subfigure}[c]{0.48\textwidth}
	\scriptsize
	\textbf{\enquote{bedroom}} (22.9\% labeled true)\\[-2mm]
  
	\parbox{\widthof{\scriptsize{W: 88.8 \textsf{\textsc{auc}}}}}{
		\raisebox{\dimexpr 6mm}{\scriptsize{E: 64.3 \textsf{\textsc{auc}}}}
	}
	\fcolorbox{white}{correct}{\includegraphics[height=7mm]{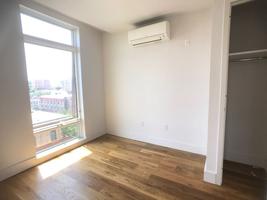}}\hfill
	\fcolorbox{white}{correct}{\includegraphics[height=7mm]{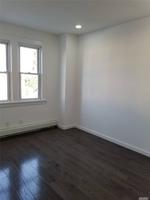}}\hfill
	\fcolorbox{white}{correct}{\includegraphics[height=7mm]{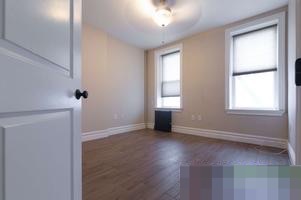}}\hfill
	\fcolorbox{white}{correct}{\includegraphics[height=7mm]{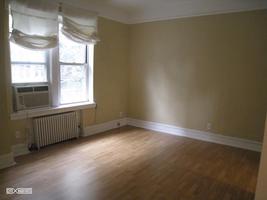}}\hfill
	\fcolorbox{white}{correct}{\includegraphics[height=7mm]{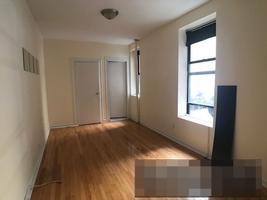}}
	\\[-2mm]

	\parbox{\widthof{\scriptsize{W: 88.8 \textsf{\textsc{auc}}}}}{
		\raisebox{\dimexpr 6mm}{\scriptsize{W: 39.0 \textsf{\textsc{auc}}}}
	}
  	\fcolorbox{white}{correct}{\includegraphics[height=7mm]{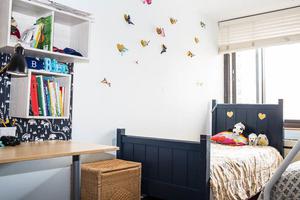}}\hfill
	\fcolorbox{white}{correct}{\includegraphics[height=7mm]{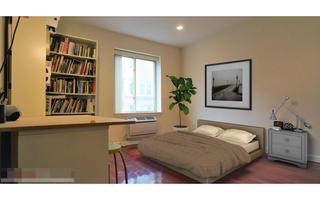}}\hfill
	\fcolorbox{white}{correct}{\includegraphics[height=7mm]{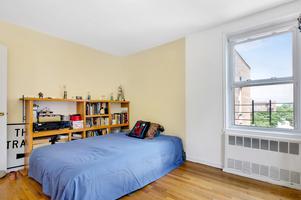}}\hfill
	\fcolorbox{white}{incorrect}{\includegraphics[height=7mm]{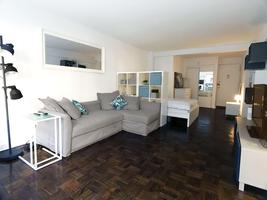}}\hfill
	\fcolorbox{white}{incorrect}{\includegraphics[height=7mm]{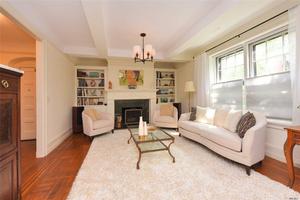}}
	\\[-2mm]

	\parbox{\widthof{\scriptsize{W: 88.8 \textsf{\textsc{auc}}}}}{
		\raisebox{\dimexpr 6mm}{\scriptsize{R: 34.4 \textsf{\textsc{auc}}}}
	}
  \fcolorbox{white}{correct}{\includegraphics[height=7mm]{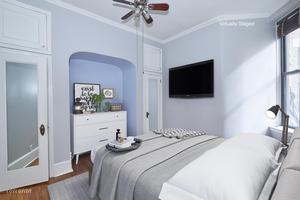}}\hfill
	\fcolorbox{white}{correct}{\includegraphics[height=7mm]{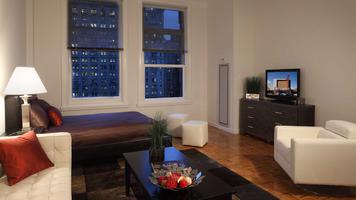}}\hfill
	\fcolorbox{white}{correct}{\includegraphics[height=7mm]{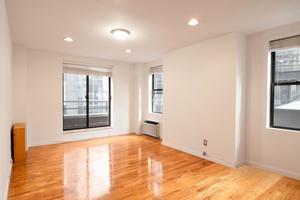}}\hfill
	\fcolorbox{white}{correct}{\includegraphics[height=7mm]{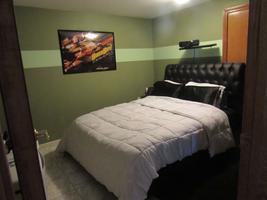}}\hfill
	\fcolorbox{white}{correct}{\includegraphics[height=7mm]{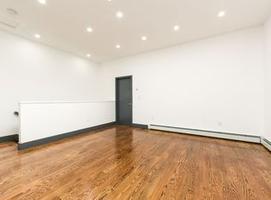}}
  \end{subfigure}
  \hfill
  \begin{subfigure}[c]{0.48\textwidth}
  	\scriptsize
	\textbf{\enquote{fitness}} (1.8\% labeled true)\\[-2mm]
  
	\parbox{\widthof{\scriptsize{W: 88.8 \textsf{\textsc{auc}}}}}{
		\raisebox{\dimexpr 6mm}{\scriptsize{E: 77.2 \textsf{\textsc{auc}}}}
	}
	\fcolorbox{white}{correct}{\includegraphics[height=7mm]{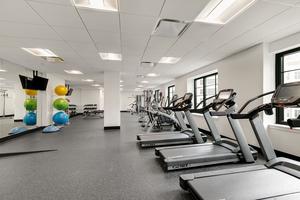}}\hfill
	\fcolorbox{white}{correct}{\includegraphics[height=7mm]{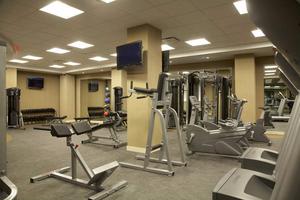}}\hfill
	\fcolorbox{white}{correct}{\includegraphics[height=7mm]{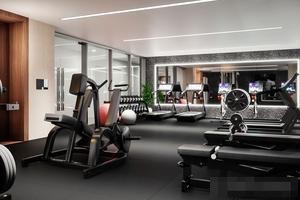}}\hfill
	\fcolorbox{white}{correct}{\includegraphics[height=7mm]{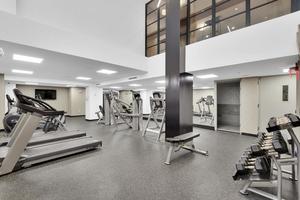}}\hfill
	\fcolorbox{white}{correct}{\includegraphics[height=7mm]{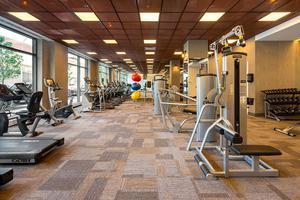}}
	\\[-2mm]

	\parbox{\widthof{\scriptsize{W: 88.8 \textsf{\textsc{auc}}}}}{
		\raisebox{\dimexpr 6mm}{\scriptsize{W: 1.5 \textsf{\textsc{auc}}}}
	}
  	\fcolorbox{white}{incorrect}{\includegraphics[height=7mm]{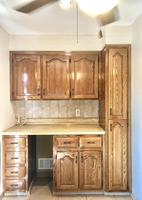}}\hfill
	\fcolorbox{white}{incorrect}{\includegraphics[height=7mm]{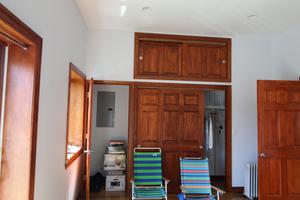}}\hfill
	\fcolorbox{white}{incorrect}{\includegraphics[height=7mm]{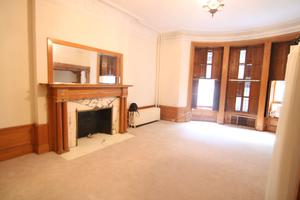}}\hfill
	\fcolorbox{white}{incorrect}{\includegraphics[height=7mm]{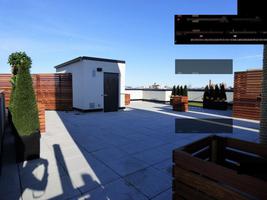}}\hfill
	\fcolorbox{white}{incorrect}{\includegraphics[height=7mm]{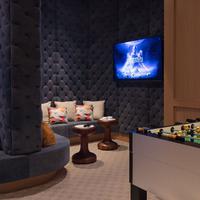}}
	\\[-2mm]

	\parbox{\widthof{\scriptsize{W: 88.8 \textsf{\textsc{auc}}}}}{
		\raisebox{\dimexpr 6mm}{\scriptsize{R: 2.4 \textsf{\textsc{auc}}}}
	}
  	\fcolorbox{white}{incorrect}{\includegraphics[height=7mm]{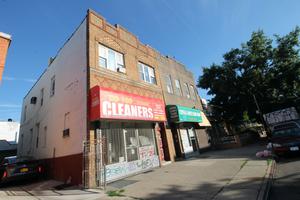}}\hfill
	\fcolorbox{white}{incorrect}{\includegraphics[height=7mm]{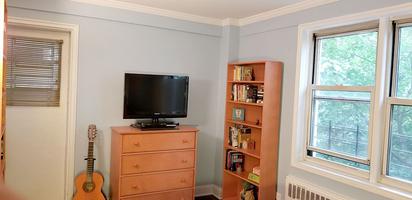}}\hfill
	\fcolorbox{white}{incorrect}{\includegraphics[height=7mm]{figures/334196396.jpg}}\hfill
	\fcolorbox{white}{correct}{\includegraphics[height=7mm]{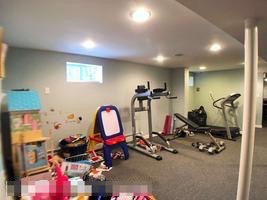}}\hfill
	\fcolorbox{white}{correct}{\includegraphics[height=7mm]{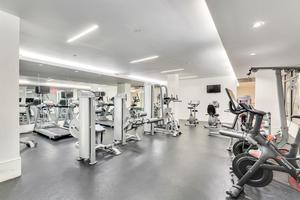}}
  \end{subfigure}

  \caption{Top images for \algname and object detection baselines on the StreetEasy dataset. {\color{cameraready} Images in each word's section come from the same evaluation set, and each row is ranked in decreasing order from left to right. For example, the three rows in the \enquote{kitchen} section are different orderings of the same 2,943 images.} Images with {\color{correct} dark blue borders} were labeled true with respect to the word, and those with {\color{incorrect} light red borders} were labeled false. E: \algname. W: word2vec object detection baseline. R: RoBERTa object detection baseline.}
  \label{fig:bestlabeled}
\end{figure*}

%% file: sections/experiments.tex
\section{Experiments}

\label{sec:sec_with_results}

\paragraph{StreetEasy dataset.}
The StreetEasy dataset comprises 29,347 real estate listings in New
York City in June 2019.  Document excerpts are shown in Figure~\ref{fig:overview}: each
consists of both images and English-language sentences.  Documents
contain an average of {\color{cameraready} 128 word tokens and} 10 images, for {\color{cameraready} totals of 3,773,608 word tokens and} 294,279 images.
There are no image-specific captions or labels. For our quantitative
word-image retrieval evaluations, we augment StreetEasy with
17,658 human relevance judgements. {\color{cameraready} After initial experiments,} we selected words with a
a variety of frequencies and degree of lexical/visual overlap with ImageNet
categories: \enquote{kitchen} (co-occurs with 200k images),
\enquote{bedroom} (175k), \enquote{washer} (65k), \enquote{outdoor}
(50k), \enquote{fitness} (49k), and \enquote{pool} (29k).  For each of
these words of interest, we labeled a different random 1\% subset of
all images (2,943 images each):
{\color{cameraready} an image in a sample was labeled true if it corresponded with any sense of the associated word and false otherwise.}
For each model, we rank images for each
query word and calculate the area under the precision-recall curve (PR
\textsf{\textsc{auc}}: perfect performance is 100, and random
performance is the percentage of images with true labels).
{\color{cameraready}
Each of the six evaluation words co-occurred with only some of their sampled images, ranging from kitchen (co-occurred with 1,997 images) to pool (310 images).
We perform evaluations on the entire samples of 2,943 images (not just those that co-occur with each word) in order to avoid overstating performance.
}


\paragraph{Experimental details for \algname.} For each image, features
are extracted from the final pre-classification layer of DenseNet169
pre-trained on ImageNet \cite{russakovsky2015imagenet} and then randomly projected from 1,664
dimensions to 256.\footnote{Random projection is a time and memory optimization. The baseline approaches have access to full feature
  vectors.}  We use a vocabulary of 7,971 words that occur at least
ten times across this corpus and Wikipedia (to eliminate
misspellings). We run \algname for 100 iterations.\footnote{The average runtime is $198 \pm 3.6$ minutes on an Intel Xeon Gold 6134 (3.20GHz) CPU with 512 GB RAM.} Setups for
baselines are comparable, and more details are available in Appendix~\ref{app:baselines}.

\paragraph{Results.}
As shown in Table~\ref{table:prauc},
\algname outperforms all baselines on PR \textsc{\textsf{auc}} on all six of the evaluation words.
The uniform initialization (Untrained \algname) is strong for frequent words (\enquote{kitchen}, \enquote{bedroom}) but poor otherwise.
The word2vec baseline is also superior to the RoBERTa baseline in {\color{cameraready} four} of six evaluations.
The baselines do best on \enquote{kitchen}, \enquote{bedroom}, and \enquote{washer}.
Table~\ref{table:objectdetection} shows the ImageNet object labels associated with each word in manually selected images.
 Though \enquote{kitchen} is not a category in the
ImageNet dataset, \enquote{microwave}, \enquote{refrigerator}, and \enquote{dishwasher} are, and these words are sufficiently close to \enquote{kitchen} to learn an association.
Nevertheless, \algname achieves the highest PR \textsc{\textsf{auc}} even in the case of \enquote{washer}, which is a category learned by the object detection baselines. 
\algname's performance increase is most pronounced for the words \enquote{outdoor},  \enquote{bedroom}, \enquote{pool}, and especially \enquote{fitness}, which have dissimilar visual manifestations in StreetEasy and ImageNet.

Qualitatively (Figure~\ref{fig:bestlabeled}), we see that \algname associates \enquote{bedroom} with empty rooms containing a door and a window while the word2vec baseline associates the word with rooms that contain a bed or a sofa.
Similarly, \enquote{outdoor} manifests in StreetEasy as building exteriors, but the RoBERTa baseline returns images of bike rooms, presumably because bicycles are usually seen outdoors.
In StreetEasy the word \enquote{pool} more frequently refers to swimming pools rather than the billiards tables seen in ImageNet.
The baseline is not technically wrong in this case (indeed, we marked pool tables as correct), but it misses the more common contextual meaning of the word in the local collection.
Finally, none of the baselines are able to handle\enquote{fitness}.

\begin{table}
\resizebox{\linewidth}{!}{%
\begin{tabular}{l c@{\hspace{.2cm}}c@{\hspace{.2cm}}c@{\hspace{.2cm}}c@{\hspace{.2cm}}c@{\hspace{.2cm}}c}
\toprule
 & washer & kitchen & outdoor & fitness & bedroom & pool\\
\hline
Random  & 1.6 & 18.4 & 16.9 & 1.8 & 22.9 & 1.3\\
word2vec & 49.3 & 52.7 & 20.0 & 1.5 & 39.0 & 20.1\\
RoBERTa & 62.1 & 21.1 & 13.2 & 2.4 & 34.4 & 17.2\\
Softmax regression & 2.0 & 19.9 & 21.6 & 3.9 & 23.0 & 13.6\\
Multinomial regression & 1.8 & 17.4 & 23.6 & 8.1 & 22.8 & 19.3\\
Untrained \algname & 1.0 & 21.6 & 10.1 & 1.4 & 42.7 & 1.2\\
\algname & \textbf{70.7} & \textbf{72.9} & \textbf{68.5} & \textbf{77.2} & \textbf{64.3} & \textbf{49.6}\\
\bottomrule
\end{tabular}
}
\caption{Area under the precision-recall curve (\textsf{\textsc{auc}}) for each grounding method on each labeled random image subset. Best-in-column is bolded. Random performance results in an \textsf{\textsc{auc}} equal to the percentage labeled true.}
\label{table:prauc}
\end{table}

{\color{cameraready}
\paragraph{Wikipedia experiments.}
We also ran \algname on a multimodal Wikipedia dataset \cite{hessel2018quantifying}.
Figure~\ref{fig:comparison} shows that the algorithm often grounds words
differently in Wikipedia's much broader range of images than it does in the StreetEasy dataset.
Similarly, top ranked images in Wikipedia for \enquote{fitness} included marathon runners rather than the StreetEasy dataset's exercise rooms.
}

%% file: sections/discussion.tex
\section{Discussion}

We present \algname, a simple clustering-based algorithm for learning image groundings for words.
{\color{cameraready} It is motivated by the unlabeled multimodal data that exists in abundance rather than relying on expensive custom datasets.}
By encouraging words to compete to claim responsibility for images, we \enquote{sharpen} the resulting image/word associations.
The method is effective at finding contextual 
{\color{cameraready} lexical groundings of words in unlabeled multi-image, multi-sentence documents}
even in the presence of  high cross-document similarity.

{\color{cameraready}
  One area for future work would be to better identify and model words that either \emph{don't} have a visual grounding or whose identified visual grounding doesn't align with human expectation. For example, the word \enquote{Gristedes} (the name of a supermarket chain) appears in StreetEasy documents, but users rarely post photographs of the supermarkets themselves. Conversely, the word
  \enquote{bright}
  outside the context of StreetEasy may not be ``visually concrete'' (according to human judgment);
  nonetheless, it frequently co-occurs with
  images of sunlit hardwood floors.
Given the lexical and visual identifiability issues explored in \S\ref{sec:sec_with_identifiability}, incorporating prior human concreteness judgments (e.g., \newcite{nelson2004university}) for vocabulary items might enable \algname to learn for these sorts of ambiguous lexical items. However, finding an appropriate balance of domain-specific flexibility versus alignment with human priors could pose a significant challenge.
}

\begin{table}
\footnotesize
\resizebox{\linewidth}{!}{%
\begin{tabular}{l c l}
\toprule
Evaluation word & Image & Top DenseNet169 predictions\\
\hline\\[-2mm]
\raisebox{\dimexpr 4mm}{\enquote{kitchen}} & \includegraphics[width=14mm]{figures/319045483.jpg} & \raisebox{\dimexpr 4mm}{`dishwasher', `microwave', `refrigerator'} \\
\raisebox{\dimexpr 4mm}{\enquote{bedroom}} & \includegraphics[width=14mm]{figures/320867106.jpg} & \raisebox{\dimexpr 4mm}{`sliding\_door', `wardrobe', `window\_shade'} \\
\raisebox{\dimexpr 4mm}{\enquote{outdoor}} & \includegraphics[width=14mm]{figures/334397895.jpg} & \raisebox{\dimexpr 4mm}{`mountain\_bike', `bicycle-built-for-two'}\\
\raisebox{\dimexpr 4mm}{\enquote{pool}} & \includegraphics[width=14mm]{figures/346500513.jpg} & \raisebox{\dimexpr 4mm}{`pool\_table', `fountain', `tub'}  \\
\raisebox{\dimexpr 4mm}{\enquote{washer}} & \includegraphics[height=10mm]{figures/350992643.jpg} & \raisebox{\dimexpr 4mm}{`washer', `microwave', `reflex\_camera'} \\
\raisebox{\dimexpr 4mm}{\enquote{fitness}} & \includegraphics[width=14mm]{figures/323191661.jpg} & \raisebox{\dimexpr 4mm}{`shoe\_shop', `dumbbell', `barbell'} \\
\bottomrule
\end{tabular}
}
\caption{Top DenseNet169 ImageNet class predictions for selected example images.}
\label{table:objectdetection}
\end{table}



%% file: sections/appendix.tex
\section{Document similarity metrics}
\label{app:similarity}

We compute a length-controlled version of word mover's distance
\cite{kusner2015word} to measure the textual distances between
documents. This was inspired by the
simple extension to ``image mover's distance'' enabled by swapping the
word2vec token representations to CNN image representations.

After computing image/word mover's distances, we noticed that
these metrics were slightly correlated with document length; this
correlation was also noted by \newcite{kusner2015word}, who mention
that longer documents might be closer to others ``as longer documents
may contain several similar words.'' To account for this, we implemented a version
of mover's distances that selects a bootstrap sample of
$b_1\!=\!50$ words and $b_2\!=\!10$ images before computing distances. The
scatterplot we report in Figure~\ref{fig:overview} is insensitive to reasonable choices of these
parameters, as it looks largely the same for any $\langle b_1, b_2
\rangle \in \{10, 30, 50 \} \times \{ 3, 5, 10 \}$.

To compute a corpus-level statistic, it's computationally infeasible to
compute distances between all possible pairs;
some calculations based on the EMD
library we are using shows that full computation would take at least a
few months. Instead, we randomly sample 10K pairs and report
confidence intervals for the mean in the figure.

\begin{figure}[b]
  \centering
    \includegraphics[width=\linewidth]{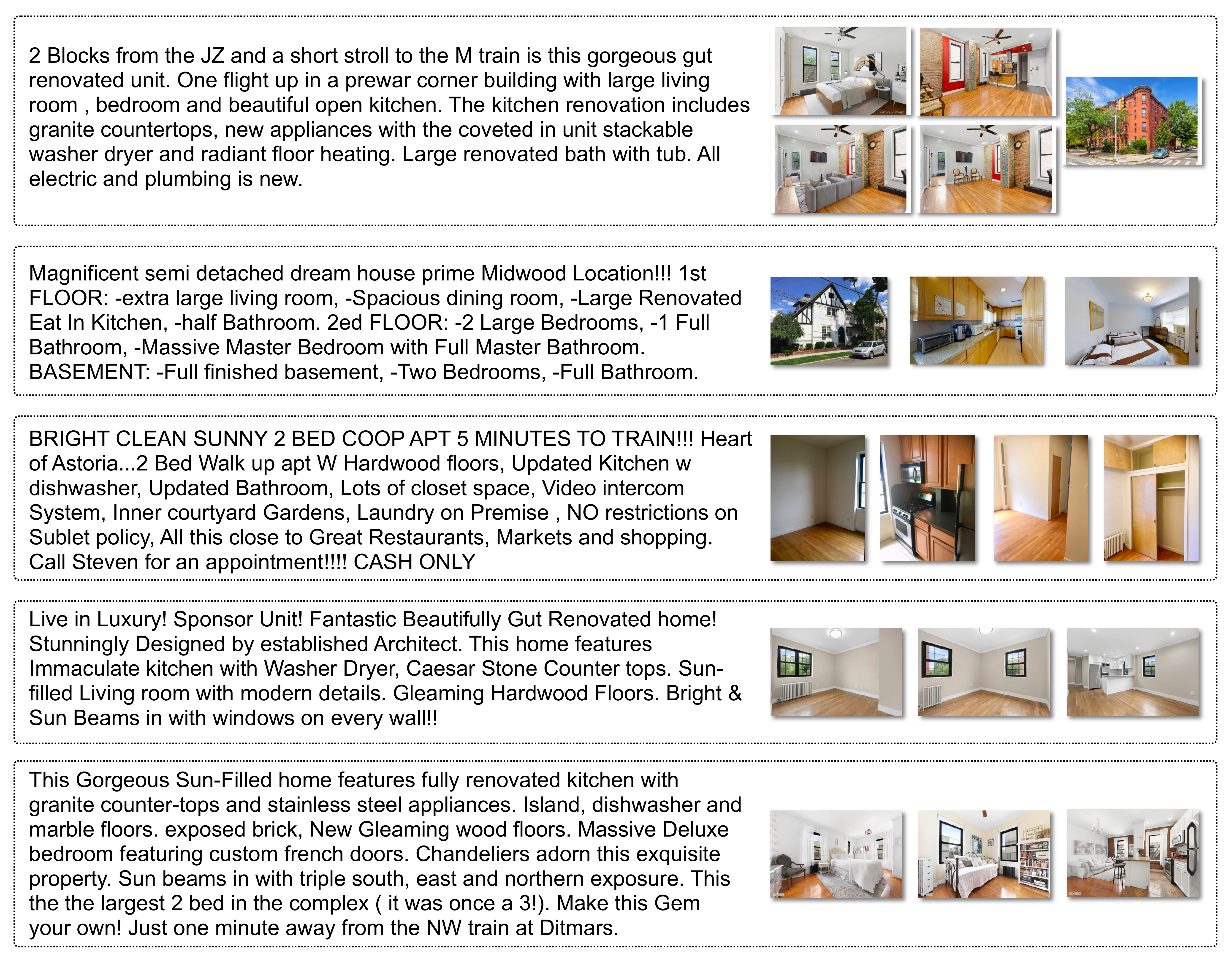}
  \caption{Additional excerpts of documents in the StreetEasy dataset.}
  \label{fig:dataset-examples}
\end{figure}

\section{StreetEasy dataset preprocessing}

The dataset consists of 29,347 English-language real estate listings from the StreetEasy website from June 2019. They contain a total of 294,279 images and 24,078,190 word tokens across 34,564 word types.
We preprocess the text by removing numbers, punctuation, hyphens, and capitalization. We restrict the vocabulary to word types that occur at least ten times in StreetEasy and in the multimodal Wikipedia dataset.
This results in 3,773,608 word tokens across 7,971 word types.
Figure~\ref{fig:overview} shows a few excerpts of listings, and
Figure~\ref{fig:dataset-examples} shows additional listing excerpts.


\section{Baselines}
\label{app:baselines}

\paragraph{Object detection.}
An image is represented as the mean of the word vectors of its top $K$ class predictions from DenseNet169. 
We report each model's performance with the $K \in \{1, \ldots, 20\}$ that resulted in the highest average PR \textsc{\textsf{auc}} across evalation words to create the strongest baselines ($K=2$ for word2vec and $K=1$ for RoBERTa). For words not in the word2vec vocabulary, we use a random vector as the word embedding. All six evaluation words are present in the word2vec vocabulary.
Average runtimes are $80.9 \pm 1.6$ seconds for word2vec and $458.8 \pm 1.6$ seconds for RoBERTa.

\paragraph{Image tagging.}

We reserved 20\% of the StreetEasy corpus as a validation set. We
don't hold out a test set: this tasks the algorithms only with
fitting the dataset, not generalizing beyond it. We use the validation
set for early stopping, model selection, and hyperparameter
optimization. We optimize learning rate (in $\{ 0.001, 0.0005, 0.0007
\}$) and number of layers (in $\{0,1,2,3,4,5\}$). We decay learning rate
upon validation loss plateau. We use the Adam optimizer
\cite{kingma2014adam}.

\section{\algname training}

\begin{figure}[b]
  \centering
    \includegraphics[width=\linewidth]{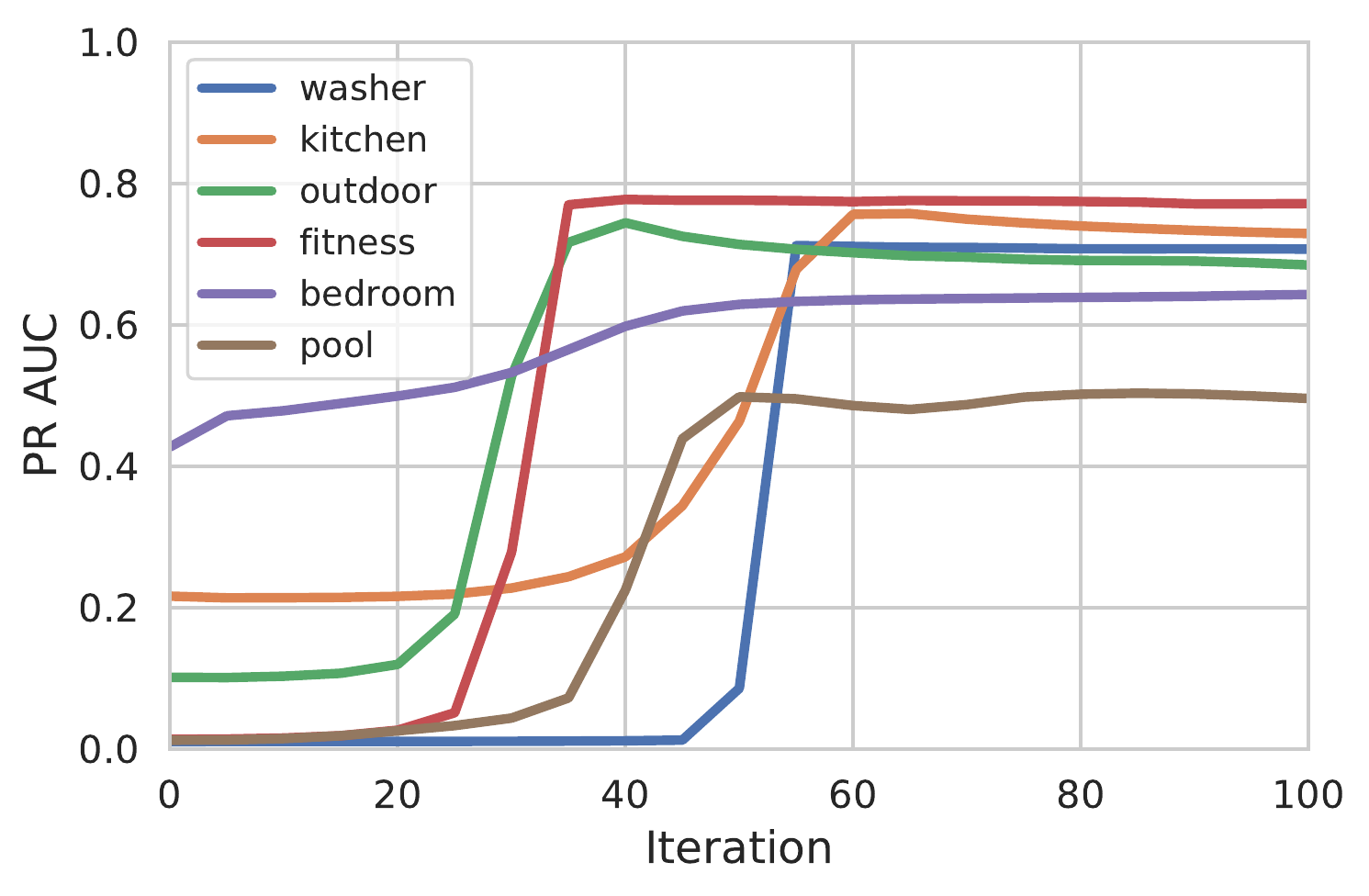}
  \caption{During \algname training, PR \textsc{\textsf{auc}} plateaus at a different rate for each evaluation word.}
  \label{fig:learning-curves}
\end{figure}

We run \algname for 100 iterations.
Figure~\ref{fig:learning-curves} shows that PR \textsc{\textsf{auc}} converges at different rates for the
different evaluation words. 

